\documentclass[conference]{IEEEtran}
\IEEEoverridecommandlockouts
\usepackage{cite}
\usepackage[hyphens]{url}
\usepackage{amsmath,amssymb,amsfonts}
\usepackage{algorithmic}
\usepackage{graphicx}
\usepackage{subfig} 
\usepackage{textcomp}
\usepackage{booktabs}
\usepackage{multirow}
\usepackage{xcolor}
\usepackage{graphicx}
\graphicspath{{images/}}            
\DeclareGraphicsExtensions{.pdf,.png,.jpg}
\def\BibTeX{{\rm B\kern-.05em{\sc i\kern-.025em b}\kern-.08em
    T\kern-.1667em\lower.7ex\hbox{E}\kern-.125emX}}
\newcommand\blfootnote[1]{%
  \begingroup
  \renewcommand\thefootnote{}\footnote{#1}%
  \addtocounter{footnote}{-1}%
  \endgroup
}

\begin{document}

\title{Diabetic Retinopathy Lesion Segmentation through Attention
Mechanisms}


\author{\IEEEauthorblockN{Aruna Jithesh, Chinmayi Karumuri, Venkata Kiran Reddy Kotha, Meghana Doddapuneni, Taehee Jeong}
\IEEEauthorblockA{\textit{San Jose State University} \\
 aruna.jithesh, chinmayi.karumuri, venkatakiranreddy.kotha, meghana.doddapuneni, taehee.jeong@sjsu.edu}
{\thanks{This work was supported in part by a Mobilint Grant awarded to San Jose State University. (Corresponding author: Taehee Jeong)}
}
}

\maketitle

\begin{abstract}
Diabetic Retinopathy (DR) is an eye disease which arises due to diabetes mellitus. It might cause vision loss and blindness. To prevent irreversible vision loss, early detection through systematic screening is crucial. Although researchers have developed numerous automated deep learning-based algorithms for DR screening, their clinical applicability remains limited, particularly in lesion segmentation. Our method provides pixel-level annotations for lesions, which practically supports Ophthalmologist to screen DR from fundus images. In this work, we segmented four types of DR-related lesions: microaneurysms, soft exudates, hard exudates, and hemorrhages on 757 images from DDR dataset. To enhance lesion segmentation, an attention mechanism was integrated with DeepLab-V3+. Compared to the baseline model, the Attention-DeepLab model increases mean average precision (mAP) from 0.3010 to 0.3326 and the mean Intersection over Union (IoU) from 0.1791 to 0.1928. The model also increased microaneurysm detection from 0.0205 to 0.0763, a clinically significant improvement. The detection of microaneurysms is the earliest visible symptom of DR. 
Our code is available at \url{https://github.com/arunajithesh123/Diabetic-Retinopathy-Lesion-Segmentation-through-Attention-Mechanisms}.

\end{abstract}


\section{Introduction}

\renewcommand{\footnoterule}{%
  \kern -3pt
  \hrule width 3in height 1pt
  \kern 2pt
}
\blfootnote{24th International Conference on Machine Learning and Applications (ICMLA), 
IEEE Copyright 2025}

\begin{figure}[ht]
  \subfloat[][]{\includegraphics[width=0.24\textwidth]{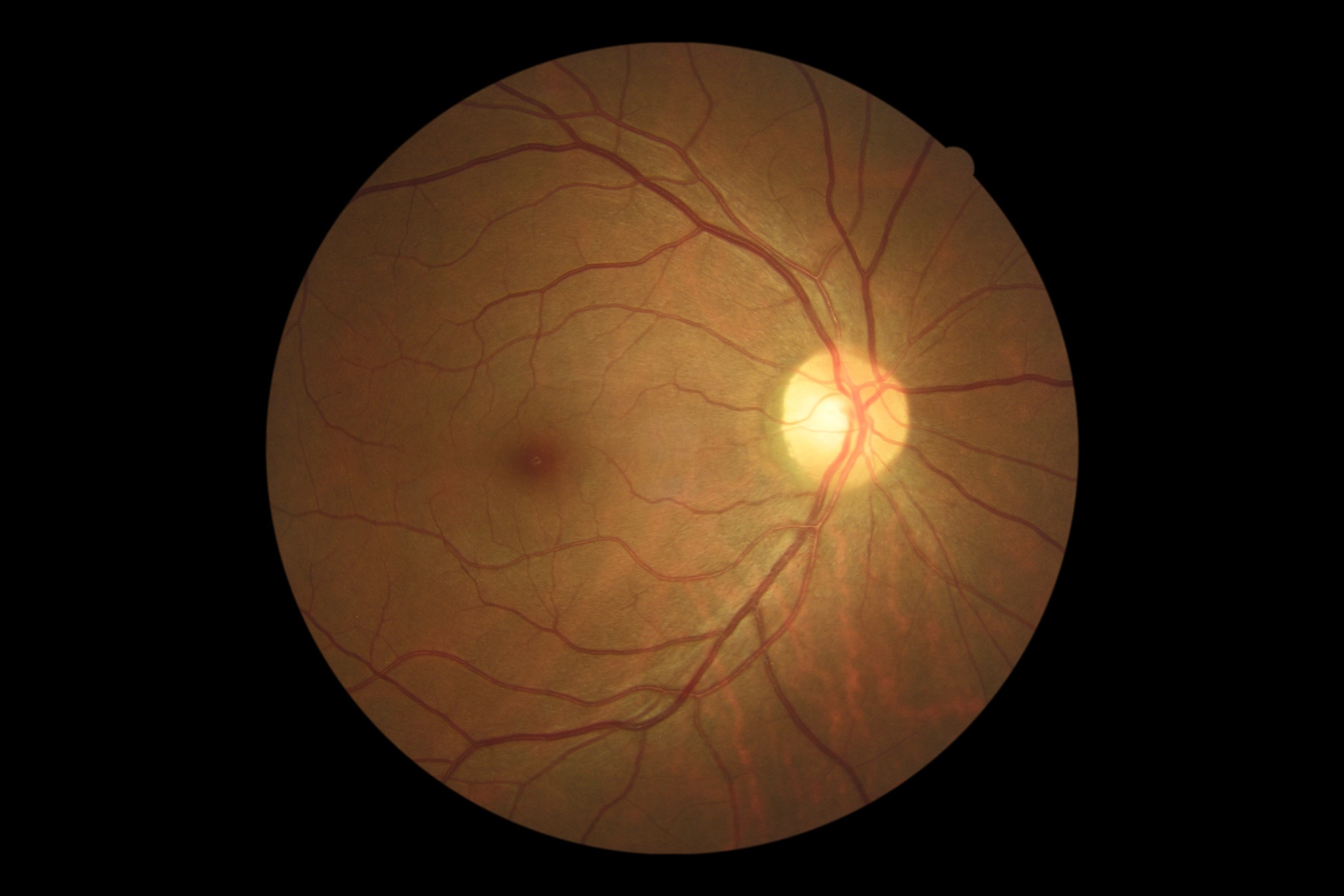}}
  \hfill
  \subfloat[][]{\includegraphics[width=0.24\textwidth]{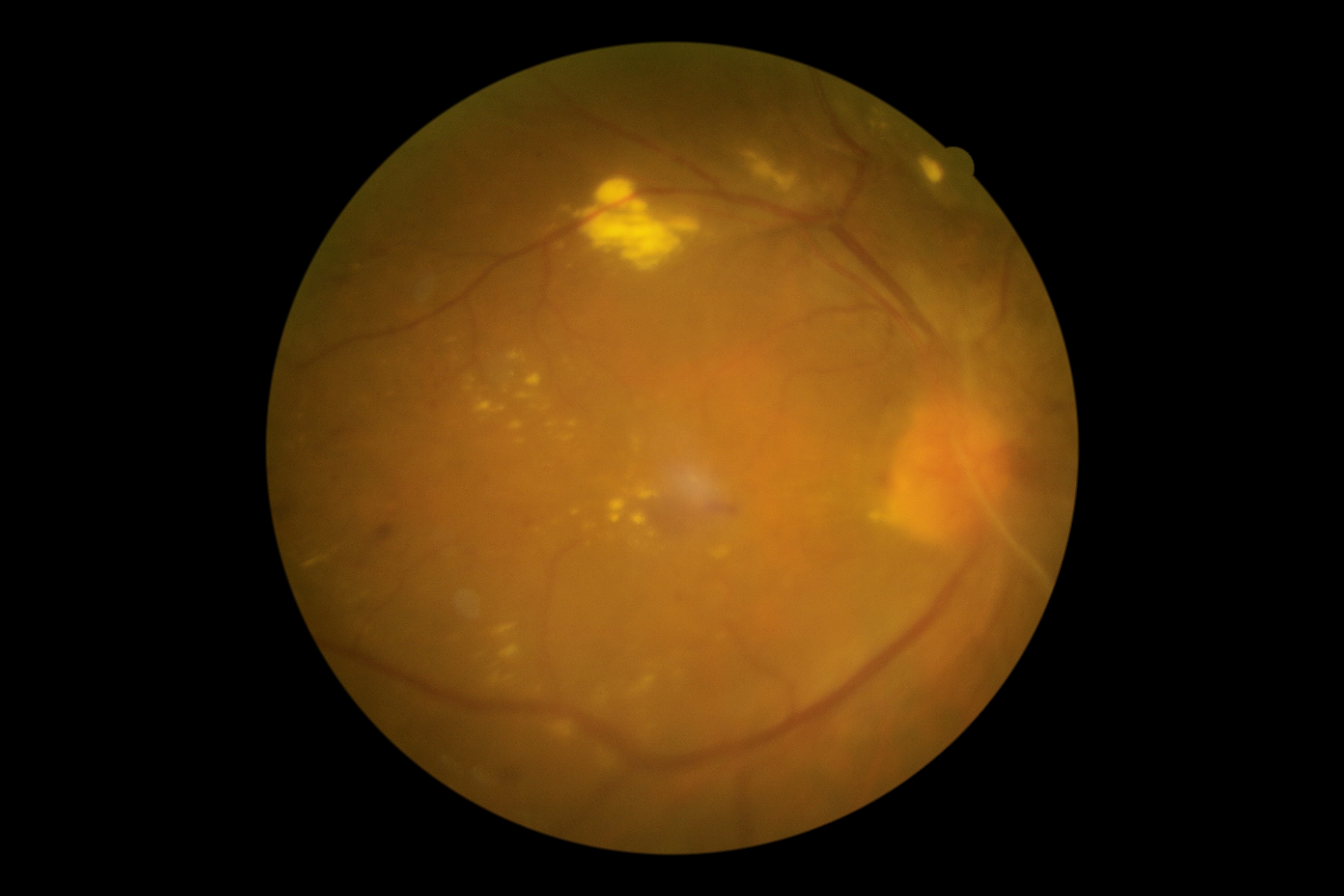}}
\caption{(a) Normal retina (b) Diabetic Retina images from DDR dataset}
\label{fig:fundus}
\end{figure}

Diabetic retinopathy (DR) is the leading cause of preventable blindness in adults. It is affecting approximately 35\% of those with diabetes and compromising the vision of over 400 million people worldwide \cite{b1}. Fig.\ref{fig:fundus} compares the Normal and Diabetic Retina.

The imperative need for effective DR screening has never been more critical, as the prevalence of diabetes is expected to reach 700 million by 2045. It is crucial to note that early detection and opportune intervention can prevent up to 90\% of severe vision loss from DR, rendering automated screening systems indispensable for the preservation of sight in this rapidly expanding patient population \cite{b2}.

Current screening programs encounter insurmountable obstacles in clinical practice, despite the preventable nature of DR-related blindness. 
The manual fundus examination of patients by trained specialists is inherently time-consuming, limiting the number of examinations performed per clinician to approximately 15-20 per day. \cite{b3}

An alternative solution to these challenges is automated DR screening, which offers several benefits, including high efficiency, low costs, and reduced reliance on clinicians \cite{b4,b5}. However, most automated DR screening algorithms provide only a DR grading result, limiting their ability to provide ophthalmologists with the evidence to make informed decisions.  Ophthalmologists typically assess the severity of DR by evaluating the presence and characteristics of lesions.  
Current models perform poorly in detecting small lesions, particularly microaneurysms, the earliest indicators of DR progression \cite{b2}.

The accuracy of microaneurysm detection is crucial for the timely implementation of interventions that prevent the progression to sight-threatening stages, as they are the earliest visible symptoms of DR. However, DeepLab-V3+\cite{b7}, the current state-of-the-art deep learning method, achieves only 2.05\% average precision for microaneurysm detection \cite{b2}.

To improve the detection of microaneurysm, this study enhances the lesion segmentation architecture by incorporating the Convolutional Block Attention Module (CBAM)\cite{b6} into the DeepLab-V3+ architecture, thereby boosting the accuracy of small, clinically significant lesions detection.


The model achieved a 272\% increase in microaneurysm detection (0.0763 vs. 0.0205 AP) and a 10.5\% increase in mean average precision across all lesion types compared to DeepLab-V3+. 
These advances represent a significant step toward clinically viable automated DR screening systems that can provide the lesion-level diagnostic information required by ophthalmologists
while maintaining the scalability needed to address the global screening crisis in diabetic eye care.

\section{Related Work}
The application of deep learning to the segmentation of diabetic retinopathy lesions has undergone a series of architectural innovations, each of which has addressed specific challenges in the detection of distinct types of retinal pathologies.

\subsection{DeepLab-V3+}
DeepLab-V3: DeepLab-V3\cite{b8} introduced Atrous Spatial Pyramid Pooling (ASPP) and Atrous (dilated) convolutions for semantic segmentation. DeepLab-V3 addressed the fundamental trade-off between spatial precision and receptive field size using dilated convolutions for multi-scale context aggregation. The ASPP module is particularly effective for segmenting objects of varying sizes, as it employs parallel Atrous convolutions with various dilation rates (6, 12, 18) to capture objects at multiple scales. The architecture exhibited state-of-the-art performance on the PASCAL VOC 2012 and Cityscapes datasets. Nevertheless, the original DeepLab-V3 was unable to accurately segment medical lesions due to the aggressive downsampling in the encoder path, which resulted in a lack of precision in the fine boundary details.

\begin{figure}[ht]
  \subfloat[][]{\includegraphics[width=0.24\textwidth]{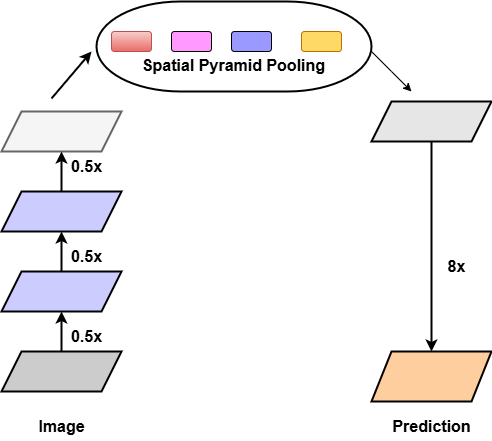}}
  \hfill
  \subfloat[][]{\includegraphics[width=0.24\textwidth]{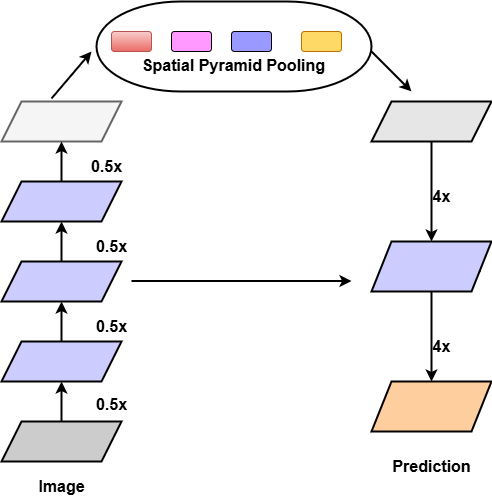}}
\caption{Neural architecture of (a) DeepLab-V3 (b) DeepLab-V3+}
\label{fig:deeplabv3+}
\end{figure}

DeepLab-V3+: Fig.\ref{fig:deeplabv3+} compares the neural architecture of and DeepLab-V3 and DeepLab-V3+.  In DeepLab-V3+, there is skip connection between low-level features in encoder and corresponding up-sampling features in decoder. It helps to recover spatial information and determine object boundaries. Although DeepLab-V3+ demonstrated exceptional performance in natural image segmentation, its application to medical imaging revealed substantial limitations, particularly in the detection of small lesions, which necessitates specialized attention mechanisms.


\subsection{Convolutional Block Attention Module}
\begin{figure*}[t]
  \centering
  \includegraphics[width=\textwidth]{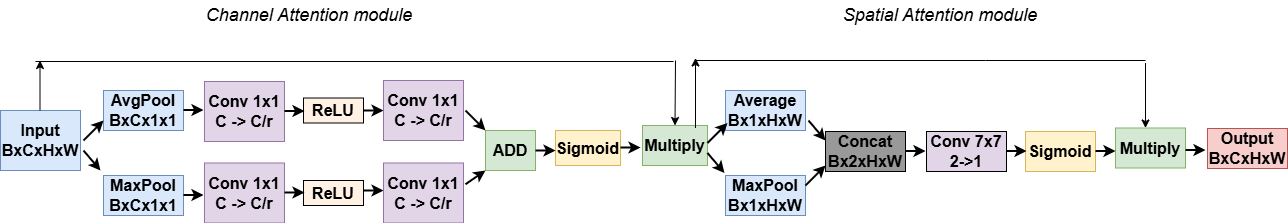}
  \caption{Convolutional Block Attention Module. 
  B: batch size, C: channel of feature map, H: height of feature map, 
  W: width of feature map, $r$: reduction ratio.}
  \label{fig:cbam}
\end{figure*}

A novel lightweight attention module,  Convolutional Block Attention Module (CBAM) was introduced, allowing for seamless integration into any existing CNN architecture without significant computational overhead. CBAM sequentially applies channel and spatial attention mechanisms to adaptively refine feature representations as illustrated in Fig.\ref{fig:cbam} . The channel attention module exploits inter-channel relationships by utilizing both average-pooled and max-pooled features through a shared multi-layer perceptron, while the spatial attention module focuses on informative spatial locations by leveraging complementary pooling operations followed by a $7 \times 7$ convolution. The sequential arrangement of channel-first, then spatial attention, proved to be the most effective across various computer vision tasks. These sub-modules function in combination to concentrate on the "what" features that are significant and the "where" to focus in the image.

\subsection{U-Net and Gated U-Net}
U\textendash Net \cite{b17} is a seminal encoder--decoder architecture for biomedical image segmentation. The encoder contracts the input with stacked $3\times3$ convolutions and $2\times2$ max-pooling (progressively widening channels, e.g., $64\!\to\!128\!\to\!256\!\to\!512\!\to\!1024$), while the decoder mirrors this path with transposed convolutions. Skip connections concatenate encoder features with decoder activations, preserving fine spatial detail and enabling precise boundary delineation. A practical limitation, however, is that these skip pathways forward encoder features indiscriminately, potentially reintroducing background clutter.

Gated U\textendash Net \cite{b18}  addresses this by inserting attention gates on skip connections to selectively pass task-relevant features. Each gate computes an attention coefficient
\begin{equation*}
  \alpha \;=\; \sigma\!\Big(\psi^{\top}\!\big(\sigma(W_{x}^{\top} x_{l} \;+\; W_{g}^{\top} g_{l} \;+\; b_{g})\big) \;+\; b_{\psi}\Big),
\end{equation*}
where $x_{l}$ is the encoder feature at level $l$ and $g_{l}$ is a coarse gating signal from the decoder. The gated feature is then
\begin{equation*}
  \hat{x}_{l} \;=\; \alpha \odot x_{l},
\end{equation*}
with $\odot$ denoting elementwise multiplication. This multiplicative gating suppresses irrelevant regions while retaining structures likely to belong to the target anatomy.

Both architectures perform strongly across medical segmentation tasks, yet in diabetic retinopathy they must contend with extreme class imbalance---lesions occupy only a tiny fraction of pixels---making small-lesion detection particularly fragile. In practice, this often necessitates complementary strategies (e.g., imbalance-aware losses or lesion-focused attention) alongside architectural improvements.

\section{Methodology}

\subsection{Dataset}
The DDR dataset\cite{b2} is made of 13,673 fundus images from 9,598 patients. For lesion segmentation tasks, a subset of 757 images were annotated with pixel-level and bounding-box. The dataset included four principal lesion types: microaneurysms (MA), hard exudates (EX), soft exudates (SE), and hemorrhages (HE). Within the 757 annotated images, MA appeared in 570 images, EX in 486, SE in 239, and HE in 601, with many images containing multiple lesion types. Each image was accompanied by four annotation masks corresponding to the lesion types, where absent lesions were represented by zero-valued masks.
Annotation files were provided in both TIFF and PNG formats, reflecting differences in annotation workflows during dataset construction. TIFF files were primarily used in earlier phases to preserve annotation fidelity without compression artifacts, whereas PNG files were adopted in later phases for alignment with standard medical imaging pipelines. For consistency, all masks were standardized to PNG format during preprocessing, with automated conversion of TIFF files when necessary. The lesion masks were subsequently stacked to form multi-channel binary masks and converted from BGR to RGB color space to ensure compatibility with pre-trained model requirements. The lesion segmentation dataset was split into train (384 images), test (226 images), and validation (150 images) sets to reduce overfitting.

\subsection{Data Preprocessing}
To enhance model generalization and address the unique challenges associated with fundus image analysis, a comprehensive augmentation strategy was developed. As a preliminary step, all images were resized to $512 \times 512$ pixels to ensure consistent input dimensions.

To account for variability in retinal image orientation due to patient positioning during acquisition, spatial transformations were applied, including horizontal and vertical flips ($p=0.5$ each), as well as random $90^\circ$ rotations ($p=0.5$). Furthermore, additional geometric augmentations, such as shift, scale, and rotation (shift limit: $\pm6.25\%$, scale limit: $\pm10\%$, rotation limit: $\pm15^\circ$, $p=0.5$), were used to simulate the variations in camera distance and alignment that can occur during image capture.

To address the variability in imaging conditions that occurs across different equipment and clinical settings, we incorporated various color augmentation techniques. These techniques, which were applied randomly through One Of selection process ($p=0.5$), included adjustments to brightness and contrast ($\pm20\%$, $p=0.4$), gamma correction (0.8–1.2, $p=0.4$), and HSV modifications (hue: $\pm10^\circ$, saturation: $\pm15\%$, value: $\pm10\%$, $p=0.4$). By applying these transformations, we aimed to simulate the differences in image appearance that can arise from factors such as varying lighting conditions, camera hardware, and patient-specific characteristics like media opacity or pigmentation.

To improve robustness to common clinical image degradations, noise augmentations were applied, including Gaussian noise (variance: 10–50, $p=0.15$) and Gaussian blur (kernel size $\leq 3$ pixels, $p=0.15$), which simulated real-world issues such as motion artifacts, focus variability, and sensor noise.

To improve the detection of small lesions, particularly microaneurysms, specialized geometric augmentations were introduced. These augmentations, which included elastic transformations ($\alpha=1$, $\sigma=50$, $p=0.15$), grid distortions ($p=0.15$), and optical distortions (distortion limit: $\pm1.0$, shift limit: $\pm0.5$, $p=0.15$), were applied using One Of selection process ($p=0.3$). This approach preserved the morphology of the lesions while simulating the natural variations that occur due to retinal curvature and changes in perspective.

To ensure compatibility with pre-trained ResNet backbones, all images were normalized using ImageNet statistics, with mean values of [0.485, 0.456, 0.406] and standard deviations of [0.229, 0.224, 0.225]. For evaluating validation and test dataset, only resizing and normalization were applied to maintain evaluation consistency. 


\subsection{DeepLab-V3+ with Attention Mechanisms}

\begin{figure}[ht]
\centerline{\includegraphics[width=0.95\linewidth]{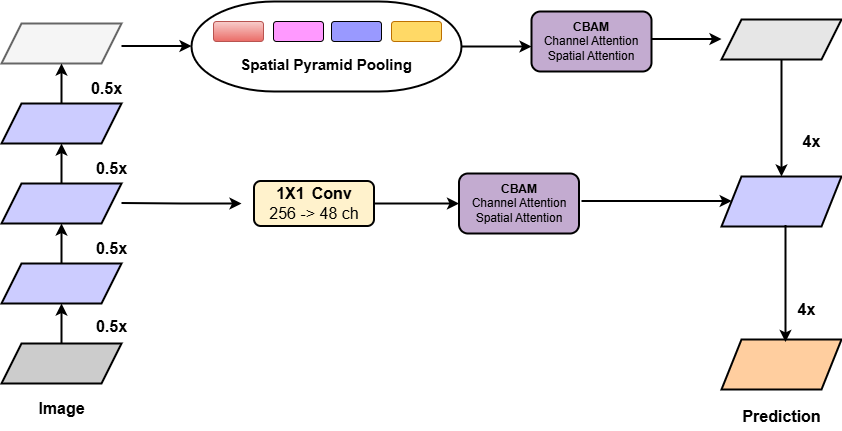}}
\caption{Neural architecture of Attention-DeepLab}
\label{fig:attentionDeeplab}
\end{figure}

This study integrates DeepLab-V3+ architecture with the CBAM module to address the challenges of small lesion detection in diabetic retinopathy screening. Fig.\ref{fig:attentionDeeplab} illustrates our neural architecture of Attention-DeepLab. CBAM modules were inserted at two critical network locations, establishing a dual-pathway attention mechanism that addresses distinct aspects of the segmentation task.  The first CBAM module is located immediately following the ASPP module. It operates on high-level semantic features with 256 channels at 16×16 resolution, highlighting lesion-specific semantic information. The 2nd CBAM module is located in the decoder pathway following the 1×1 convolution that reduces low-level features from 256 to 48 channels at 128×128 resolution. It enhances to refine spatial details and precise boundary information that are essential for accurate lesion delineation.

 A hierarchical dual-attention mechanism is established by this novel attention-enhanced architecture, which represents a substantial improvement over the baseline DeepLab-V3+. High-level features receive semantic-focused attention, while low-level features receive spatially-focused attention.
 The architecture is particularly effective in detecting microaneurysms, which are the earliest indicators of DR progression but present the greatest detection challenge due to their minimal pixel imprint.




\subsection{Loss Function}

The binary cross-entropy loss is used for binary classification task. It quantifies the difference between predicted probabilities and actual binary labels, penalizing inaccurate predictions. 

\begin{equation}
\mathcal{L}_{\text{BCE}} = -\frac{1}{N} \sum_{i=1}^{N} \left[\cdot y_i \log(\sigma(x_i)) + (1 - y_i) \log(1 - \sigma(x_i)) \right]
\end{equation}

where $N$ is the number of pixels, $y_i \in \{0, 1\}$ is the ground truth label, $x_i$ is the model output, $\sigma(\cdot)$ is the sigmoid function.

The focal loss\cite{b9} tackles class imbalance by emphasizing hard-to-classify examples and reducing the weight of easier ones:

\begin{equation}
\mathcal{L}_{\text{Focal}} = -\alpha_t (1 - p_t)^\gamma \log(p_t)
\end{equation}

\noindent
Where:
\begin{itemize}
    \item $p_t = p$ if $y = 1$, or $p_t = 1 - p$ if $y = 0$
    \item $p = \text{sigmoid}(\text{output})$ (predicted probability)
    \item $\alpha = 0.25$ (balancing factor for positive/negative classes)
    \item $\gamma = 2.0$ (focusing parameter – higher values focus more on hard examples)
\end{itemize}

The Dice loss\cite{b10,b11} is based on the Dice similarity coefficient. It is commonly used in medical image segmentation to tackle class imbalance between foreground and background pixels. The Dice coefficient measures how much predicted segmentations overlap with the ground truth segmentations:

\begin{equation}
\text{Dice} = \frac{2|P \cap T| + \epsilon}{|P| + |T| + \epsilon}
\end{equation}

where $P$ represents the predicted segmentation, $T$ denotes the ground truth, $|P \cap T|$ is the intersection, and $\epsilon$ is a smoothing factor to prevent division by zero. The Dice loss is then defined as:

\begin{equation}
\mathcal{L}_{\text{Dice}} = 1 - \text{Dice}
\end{equation}

The boundary loss\cite{b12,b13} highlights precise marking of lesion boundaries by using spatial gradient information:

\begin{equation}
\mathcal{L}_{\text{Boundary}} = \mathcal{L}_{\text{BCE}} \cdot \exp\left(-\theta \cdot \|\nabla T\|\right)
\end{equation}

\noindent
where $\|\nabla T\|$ represents the gradient magnitude of the ground truth mask computed using Sobel operators:

\begin{equation}
\|\nabla T\| = \sqrt{(\nabla_x T)^2 + (\nabla_y T)^2 + \epsilon}
\end{equation}

\noindent
The parameter $\theta$ controls how much focus is given to boundary areas. Regions with higher gradient magnitudes, which indicate closeness to boundaries, receive more attention in training.

This study implements a combined loss function that mitigates the obstacles of boundary precision and class imbalance:
\begin{align}
L &= w_{\text{dice}} \cdot L_{\text{dice}} + w_{\text{bce}} \cdot L_{\text{bce}} \notag \\
  &\quad + w_{\text{focal}} \cdot L_{\text{focal}} + w_{\text{boundary}} \cdot L_{\text{boundary}} \tag{1}
\end{align}

where $L_{\text{dice}}$ is the Dice loss, $L_{\text{bce}}$ is the binary cross-entropy loss, $L_{\text{focal}}$ is the focal loss with parameters $\alpha=0.25$ and $\gamma=2.0$, and $L_{\text{boundary}}$ is a specialized boundary loss with parameter $\theta=1.5$ for precise lesion boundary detection. Based on empirical evaluation, the weights $w_{\text{dice}}$, $w_{\text{bce}}$, $w_{\text{focal}}$, and $w_{\text{boundary}}$ are assigned to 1.0, 0.5, 1.0, and 0.5, respectively.


\subsection{Implementation Details}

PyTorch was employed to conduct model training on Nvidia A100 GPU. The Adam optimizer ($\beta_1=0.9$, $\beta_2=0.999$) was employed for lesion segmentation tasks, with a learning rate of $1 \times 10^{-4}$, which was regulated by the ReduceLROnPlateau scheduler. 
To address GPU memory limitations, we utilized a batch size of 4 and implemented early stopping after 15 epochs.

\subsection{Evaluation Metrics}

Intersection over Union (IoU)\cite{b14,b15} measures the overlap between predicted and actual regions. IoU values range from 0 (no overlap) to 1 (perfect overlap).

\begin{equation}
\text{IoU} = \frac{|P \cap T| + \epsilon}{|P \cup T| + \epsilon} = \frac{|P \cap T| + \epsilon}{|P| + |T| - |P \cap T| + \epsilon}
\end{equation}

\noindent
where $|P \cup T|$ represents the union of predicted and ground truth sets. 

\medskip

Precision is defined as $P = \frac{TP}{TP + FP}$, which is the proportion of predicted lesions that are correct. Average Precision (AP)\cite{b16} is mean of Precision at various IoU from 0.50 to 0.95.  
AP is especially helpful for assessing detection and segmentation tasks with class imbalance. It gives a single score that summarizes performance across all confidence thresholds. Mean Average Precision (mAP) calculates the average precision across multiple classes and then takes the mean of those average precisions.

\begin{equation}
\text{mAP} = = \frac{1}{N}{\sum_{i=1}^{N} (AP_{i})}
\end{equation}

The performance of lesion segmentation was assessed using 
IoU, and AP to assure a comprehensive evaluation of boundary delineation and detection accuracy.

\section{Experimental Results}

\begin{figure}[ht]
\begin{center}
  \subfloat[][Input]{\includegraphics[width=0.32\linewidth]{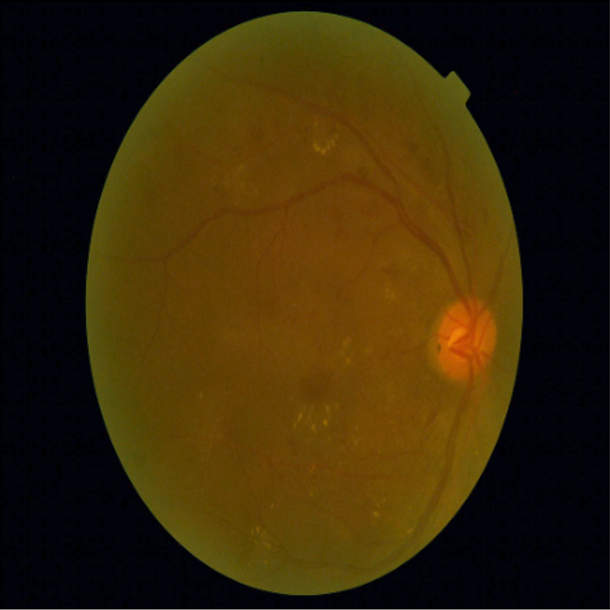}}
  \hfil
  \subfloat[][Ground Truth]{\includegraphics[width=0.32\linewidth]{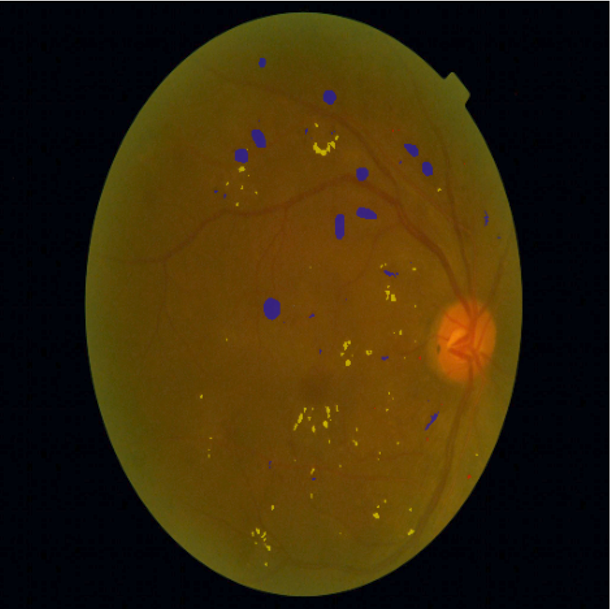}}
  \hfil
  \subfloat[][Prediction]{\includegraphics[width=0.32\linewidth]{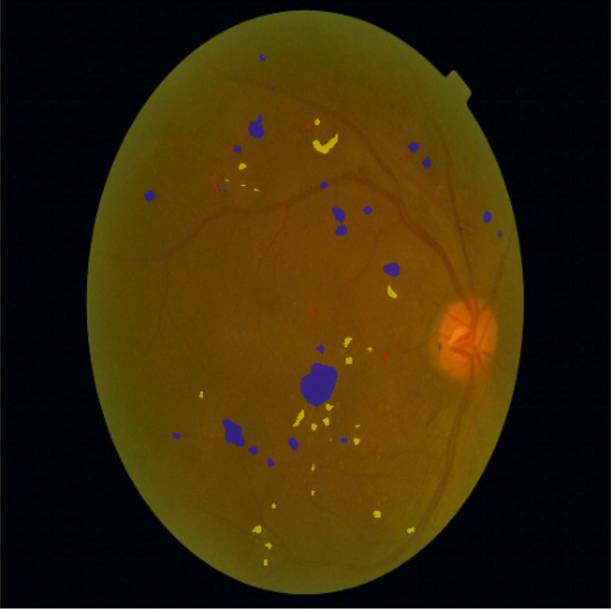}}

\end{center}
\caption{Segmentation example 1: (a) Input image, (b) Ground Truth, and (c) Prediction} 
\label{fig:seg1}
\end{figure}

\begin{figure}[ht]
\begin{center}
  \subfloat[][Input]{\includegraphics[width=0.32\linewidth]{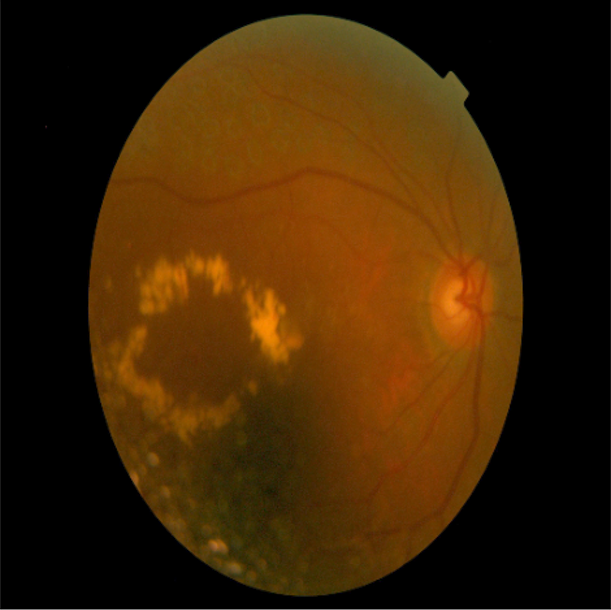}}
    \hfil
  \subfloat[][Ground Truth]{\includegraphics[width=0.32\linewidth]{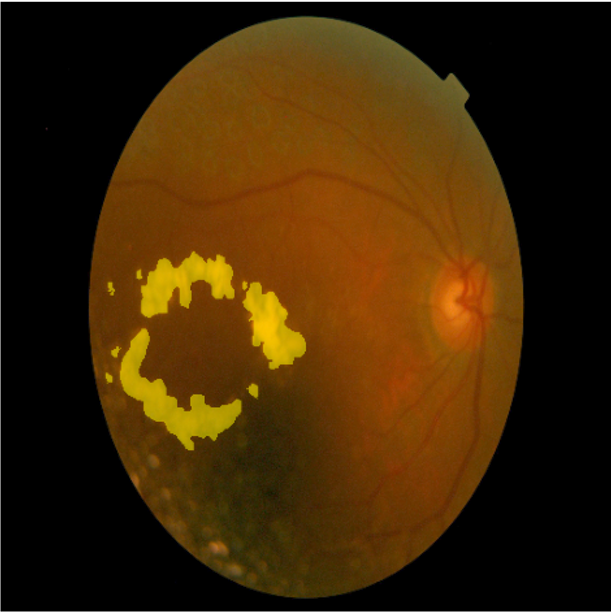}}
    \hfil
  \subfloat[][Prediction]{\includegraphics[width=0.32\linewidth]{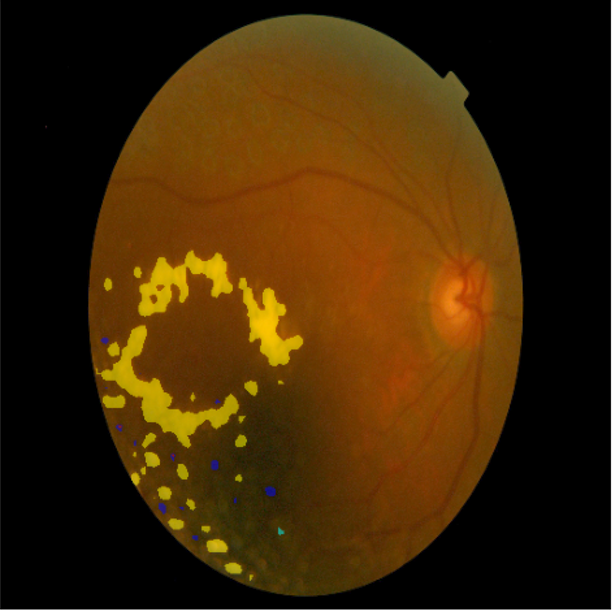}}

\end{center}
\caption{Segmentation example 2: (a) Input image, (b) Ground Truth, and (c) Prediction} 
\label{fig:seg2}
\end{figure}

Visual comparison of Lesion segmentation results showing Original images, Ground truth, and predictions from our model are illustrated in Fig. \ref{fig:seg1} and Fig. \ref{fig:seg2}, where green marks represent EX, red marks represent MA, blue marks represent HE, and light blue marks represent SE. The performance comparison between DeepLab-V3+ and our attention-based model for lesion segmentation on the test set is shown in Table \ref{tab_AP}.

Across all lesion types, DeepLab\textendash V3+ achieved $\mathrm{mAP}=0.3010$ and $\mathrm{mIoU}=0.1791$, while U\textendash Net achieved $\mathrm{mAP}=0.2391$ and $\mathrm{mIoU}=0.1757$. The proposed Attention\textendash DeepLab improved overall segmentation with $\mathrm{mAP}=0.3326$ and $\mathrm{mIoU}=0.1928$. Lesion-wise, microaneurysms (MA) and hemorrhages (HE) show the largest gains: MA improves by $+272\%$ over DeepLab\textendash V3+ ($0.0763$ vs.\ $0.0205$) and $+60\%$ over U\textendash Net ($0.0763$ vs.\ $0.0476$); HE improves by $+134\%$ over DeepLab\textendash V3+ ($0.4308$ vs.\ $0.1842$) and $+14\%$ over U\textendash Net ($0.4308$ vs.\ $0.3784$). For hard exudates (EX), Attention\textendash DeepLab is slightly below DeepLab\textendash V3+ but outperforms U\textendash Net across most lesion types.


\begin{table}[ht]
\centering
\caption{Comparison of AP and IoU scores for different lesion types across models}
\label{tab_AP} 
\begin{tabular}{lccc}
\toprule
\textbf{Metric} & \textbf{DeepLab-V3+} & \textbf{UNet} & \textbf{Attention-DeepLab} \\
\midrule
AP (MA) & 0.0205 & 0.0476 & 0.0763 \\
AP (EX) & 0.5634 & 0.3578 & 0.3960 \\
AP (SE) & 0.4359 & 0.1725 & 0.4271 \\
AP (HE) & 0.1842 & 0.3784 & 0.4308 \\
\textbf{mAP} & \textbf{0.3010} & \textbf{0.2391} & \textbf{0.3326} \\
\midrule
IoU (MA) & 0.0325 & 0.0472 & 0.0717 \\
IoU (EX) & 0.3118 & 0.2790 & 0.2742 \\
IoU (SE) & 0.2295 & 0.1214 & 0.1420 \\
IoU (HE) & 0.1425 & 0.2550 & 0.2833 \\
\textbf{mIoU} & \textbf{0.1791} & \textbf{0.1757} & \textbf{0.1928} \\
\bottomrule
\end{tabular}
\end{table}

\section{Discussion}
The critical limitation of the previous method is addressed by the significant enhancements in microaneurysm detection (272\%). The earliest visible symptom of diabetic retinopathy is microaneurysms, which are difficult to detect due to their small pixel presence in retinal images. The model effectively focused on these small but clinically significant lesions by the attention mechanisms, which may facilitate earlier diagnosis and intervention.

This study was evaluated on a single dataset (DDR, 757 images), which constrains the ability to generalize findings across diverse patient populations, imaging equipment, and clinical settings. The relatively modest training set of 384 images further limits the model’s exposure to variability in lesion morphology and acquisition conditions. Nevertheless, the DDR dataset offers high-quality pixel-level annotations, a rarity in medical imaging, which enables precise and reliable assessment of segmentation performance. Its size is comparable to that used in other recent DR lesion segmentation studies, supporting the validity of our experimental framework. The achieved improvements over established baselines within this context highlight the method’s potential. However, without external validation on independent datasets, cross-domain robustness remains unverified, and the lack of demographic characterization in DDR may further restrict broader clinical applicability.

The differential performance across lesion types reflects both the characteristics of the proposed dual-pathway attention mechanism and the underlying lesion morphology. Small lesions such as microaneurysms (MA) and hemorrhages (HE) benefited substantially from the enhanced spatial attention, which preserved fine-grained details essential for detection in noisy retinal backgrounds. Notably, microaneurysm detection improved by 272\%, underscoring the effectiveness of the approach for clinically critical early-stage lesions.  

In contrast, performance on hard exudates declined by approximately 30\%. This reduction is likely attributable to the attention mechanism fragmenting larger, coherent lesion regions by over-emphasizing local boundary refinements at the expense of global shape consistency. Hard exudates typically appear as well-defined, high-contrast regions where global contextual information contributes more to accurate delineation than fine-scale feature enhancement.  

This trade-off aligns with clinical priorities. Microaneurysms serve as the earliest detectable indicators of DR and require prompt intervention, while hard exudates are visually apparent to clinicians and generally manifest at later stages of disease progression. Thus, the overall positive impact on detection capability---reflected in the increase in mean average precision from 0.3010 to 0.3326---demonstrates the clinical value of the proposed method despite the observed decline in exudate segmentation.

Numerous clinical implications are associated with the improved lesion segmentation techniques. The most effective intervention for diabetic retinopathy is achieved through the early detection of microaneurysms and precise lesion localization at the earliest stage. The exceptional performance of the proposed model in automated lesion segmentation could facilitate dependable screening protocols, allowing specialists to focus on confirmed cases with significant lesion burdens that require immediate intervention.

The diagnostic process is rendered more transparent and reliable for clinical use by the system's provision of quantitative measurements and detailed lesion maps, which facilitate more informed treatment decisions. The robust performance in lesion detection and segmentation suggests that it is suitable for deployment in tele-medicine scenarios. This has the potential to increase access to DR screening and monitoring in underserved areas where specialist ophthalmologists are limited.

There are still a multitude of limitations, despite the significant progress that has been made. The present methodology evaluates individual images without taking into account temporal disease progression patterns, and the model's performance may be compromised by poor quality fundus images or the diverse illumination conditions that are frequently encountered in clinical practice. Scalability to new datasets without substantial labeling efforts is also restricted by the dependence on extensive pixel-level annotations.  Future research should prioritize external validation of disparate datasets from diverse populations and fundus imaging equipment. Also, it is necessary to deploy model on mobile devices and point-of-care systems.

\section{Future Work}
While the proposed Attention-DeepLab model demonstrates substantial improvements in detecting diabetic retinopathy lesions, several promising directions remain for future exploration and enhancement.

Vision transformers have shown strong potential in medical image analysis due to their ability to model long-range dependencies and global contextual information. Future work will investigate transformer-based attention mechanisms and hybrid CNN--Transformer architectures to better capture spatial relationships among lesions across fundus images.

Although CBAM provides effective channel and spatial attention, alternative mechanisms such as self-attention and multi-head attention may further improve feature discrimination. These approaches could enable the model to simultaneously attend to diverse lesion characteristics and dynamically adapt to different lesion types.

The current evaluation is limited to the DDR dataset. To assess robustness and real-world applicability, future studies will include external validation on datasets collected from multiple institutions, imaging devices, and patient populations.

An important extension is the analysis of longitudinal fundus images to track lesion evolution over time. Modeling disease progression could enable early risk stratification and more personalized intervention strategies.

Pixel-level annotation is labor-intensive and costly. Future research will explore semi-supervised, weakly supervised, and few-shot learning paradigms to reduce labeling requirements while maintaining segmentation performance.

Clinical fundus images often suffer from illumination variation, blur, and sensor noise. Incorporating adaptive preprocessing and robustness-enhancing training strategies will improve model reliability, particularly in resource-constrained clinical environments.

Ultimately, clinical validation through prospective studies comparing model performance against expert ophthalmologists is necessary. Additionally, optimizing inference efficiency for deployment on mobile and point-of-care devices will facilitate practical clinical adoption.

\section{Conclusion}

This study substantially improves the automated assessment of DR through attention mechanisms for lesion segmentation. Compared to the DeepLab-V3+, the Attention-DeepLab achieved a 10.5\% increase in mAP and a 7.6\% improvement in mIoU. Especially notable are the improvements in small lesion detection, including a 272\% improvement for microaneurysms (MA) and a 134\% improvement for hemorrhages (HE).

These technical advancements result in significant clinical benefits, such as improved accuracy for detecting challenging small lesions, enhanced early detection capabilities for sight-threatening conditions, and significant progress toward reliable automated DR screening systems based on precise lesion quantification. The limitations identified in previous studies regarding small lesion detection and segmentation accuracy are addressed in this work, which represents a substantial advancement at the intersection of research and clinical implementation of automated diabetic retinopathy lesion segmentation technology.

\vspace{12pt}
\color{red}


\begin{thebibliography}{00}
\bibitem{b1} D. S. W Ting, C. Y-L Cheung, and Gilbert Lim, ``Development and Validation of a Deep Learning System for Diabetic Retinopathy and Related Eye Diseases Using Retinal Images From Multiethnic Populations With Diabetes,'' JAMA, vol. 318, no. 22, pp. 2211-2223, 2017.

\bibitem{b2} T. Li, Y. Gao, K. Wang, S. Guo, H. Liu, and H. Kang, "Diagnostic assessment of deep learning algorithms for diabetic retinopathy screening," Information Sciences, vol. 501, pp. 511-522, 2019.

\bibitem{b3} V. Gulshan, L. Peng, and Marc Coram, ``Development and validation of a deep learning algorithm for detection of diabetic retinopathy in retinal fundus photographs,'' JAMA, vol. 316, no. 22, pp. 2402-2410, 2016.


\bibitem{b4} R. Gargeya and T. Leng, ``Automated identification of diabetic retinopathy using deep learning,`` Ophthalmology, vol. 124, no. 7, pp. 962-969, 2017.

\bibitem{b5} G. Quellec, K. Charrière, Y. Boudi, B. Cochener, and M. Lamard, ``Deep image mining for diabetic retinopathy screening,`` Medical Image Analysis, vol. 39, pp. 178-193, 2017.

\bibitem{b6} S. Woo, J. Park, J. Y. Lee, and I. S. Kweon, ``CBAM: Convolutional block attention module,`` in Proceedings of the European conference on computer vision (ECCV), 2018, pp. 3-19.

\bibitem{b7} L.-C. Chen, Y. Zhu, G. Papandreou, F. Schroff, and H. Adam, ``Encoder-decoder with atrous separable convolution for semantic image segmentation,`` in Proceedings of the European Conference on Computer Vision (ECCV), 2018, pp. 801-818.


\bibitem{b8} L.-C. Chen, G. Papandreou, F. Schroff, and H. Adam, ``Rethinking atrous convolution for semantic image segmentation,`` arXiv:1706.05587, 2017.


\bibitem{b9} T. Y. Lin, P. Goyal, R. Girshick, K. He, and P. Dollár, ``Focal loss for dense object detection,`` Proc. IEEE Int. Conf. Comput. Vis. (ICCV), 2017, pp. 2980--2988.

\bibitem{b10} F. Milletari, N. Navab, and S. A. Ahmadi, ``V-Net: Fully convolutional neural networks for volumetric medical image segmentation,`` Proc. 4th Int. Conf. 3D Vision (3DV), 2016, pp. 565--571.

\bibitem{b11} C. H. Sudre, W. Li, T. Vercauteren, S. Ourselin, and M. J. Cardoso, ``Generalised Dice overlap as a deep learning loss function for highly unbalanced segmentations,`` Deep Learn. Med. Image Anal. Multimodal Learn. Clin. Decis. Support, Springer, 2017, pp. 240--248.

\bibitem{b12} H. Kervadec, J. Bouchtiba, C. Desrosiers, E. Granger, J. Dolz, and I. B. Ayed, ``Boundary loss for highly unbalanced segmentation,`` Proc. Med. Imaging Deep Learn. (MIDL), 2019, pp. 285--296.

\bibitem{b13} F. Caliva, C. Iriondo, A. M. Martinez, S. Majumdar, and V. Pedoia, ``Distance map loss penalty term for semantic segmentation,`` arXiv:1908.03679, 2019.

\bibitem{b14} P. Jaccard, ``The distribution of the flora in the alpine zone,`` New Phytol., vol. 11, no. 2, pp. 37--50, 1912.

\bibitem{b15} M. A. Rahman and Y. Wang, ``Optimizing intersection-over-union in deep neural networks for image segmentation,`` Proc. Int. Symp. Visual Comput. (ISVC), 2016, pp. 234--244.

\bibitem{b16} J. Redmon and A. Farhadi, ``YOLOv3: An Incremental Improvement,`` arXiv:1804.02767, 2018.

\bibitem{b17}
O. Ronneberger, P. Fischer, and T. Brox, ``U-Net: Convolutional Networks for Biomedical Image Segmentation,`` in Medical Image Computing and Computer-Assisted Intervention (MICCAI), 2015, pp. 234--241.

\bibitem{b18}
O. Oktay et al., ``Attention U-Net: Learning Where to Look for the Pancreas,`` in Medical Imaging with Deep Learning (MIDL), 2018, pp. 1--10.



\end{thebibliography}
\end{document}